# "Conditional Inter-Causally Independent" node distributions, a property of "noisy-or" models

**John Mark Agosta**
Robotics Laboratory
Stanford University
Stanford, CA 94305
johnmark@flamingo.stanford.EDU

## Abstract

This paper examines the interdependence generated between two parent nodes with a common instantiated child node, such as two hypotheses sharing common evidence. The relation so generated has been termed "inter-causal." It is shown by construction that inter-causal independence is possible for binary distributions at one state of evidence. For such "CICI" distributions, the two measures of inter-causal effect, "multiplicative synergy" and "additive synergy" are equal. The well known "noisy-or" model is an example of such a distribution. This introduces novel semantics for the noisy-or, as a model of the degree of conflict among competing hypotheses of a common observation.

In a general Bayesian network, the relation between a pair of nodes can be *predictive*, meaning we are interested in the effect of a node upon its successors, or, oppositely, *diagnostic*, where we infer the state of a node from knowledge of its successors. We can define yet a third relation between nodes that are neither successors of each other, but share a common successor. Such a relation has been termed *inter-causal*. [Henrion and Druzel 1990, p.10] For example, in the simplest diagram with this property, nodes $A$ and $B$ in Figure one are inter-causally related to each other by their common evidence at node $e$. This relation is a property of the clique formed by "marrying the parents" of $e$, not by the individual effects of the arcs into $e$. In this paper I derive the quantitative inter-causal properties due to evidence nodes constructed from the noisy-or" model.

The interest in inter-causal relations occurs in the process of *abduction*, that is, reasoning from evidence back to the hypotheses that explain the evidence. This arises in problems of interpretation, where more than one hypothesis may be suggested by a piece of evidence. [Goldman and Charniak 1990] Having multiple explanations denotes the ambiguity due to not having enough information to entirely resolve which hypothesis offers the true explanation. This paper shows how to construct an evidence node that expresses this ambiguity by the degree of conflict between hypotheses. We apply this elsewhere [Agosta 1991] as a component in building a "recognition network" where relevant hypotheses are created "on the fly" as possible interpretations of the evidence.

The implicit relation between $A$ and $B$ due to shared evidence has been extensively explored as the property of one hypothesis to "explain away" another. These are cases where, given evidence and the assertion of one hypothesis, the other hypothesis can be disqualified as a cause of the evidence. This paper explores how this dependency induced between hypotheses changes with the evidence. Interestingly, with binary variables, the induced dependency may vary, and as shown by the noisy-or, disappear for certain states of evidence.

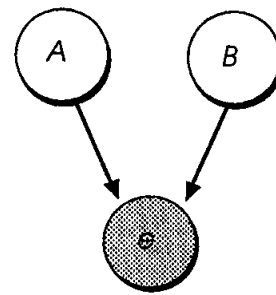

Figure 1: The relationship between hypotheses is determined by their common evidence

## 1 EVIDENCE NODES THAT ARE COMMON TO MULTIPLE PARENTS

This paper characterizes quantitatively the dependency between $A$ and $B$ that stems from the likelihood matrix at $e$. Capital letters such as $A$ and $B$ denote unobserved random variables and lower case letters denote variables when they have been observed: $e^+$ for $E = true$ and $e^-$ for $E = false$.

Dependencies between two hypotheses' existence can occur in two senses: they *conflict*, so as the probability of one hypothesis' existence increases, the other decreases—we say one tends to *exclude* the other; or, as one increases the other increases also. The latter relation shall be called *collaboration*. First I discuss some of the basic independence properties of the network shown in figure one as it depends on the state of node e. Next I consider how the conditional distribution of node e leads to conditional dependence of its parents, using the "noisy-or" model as an example for node $e$. Finally I propose a quantitative parameterization of the dependence generated between the parent nodes.

## 1.1 INTER-CAUSAL INDEPENDENCE

The definition of d-separation [Pearl 1988, p.117] provides general conditions about the conditional independence of nodes that are parents of a common evidence node. In figure one, nodes $A$ and $B$ must be independent when their common successor is uninstantiated, or has any instantiated successors. The converse is not always true: it is possible to construct cases where $A$ and $B$ remain conditionally independent after $e$ has been observed.[1] The d-separation theorem applies to the structure of the network: this conditional case extends it to the property of the distributions for a common successor node.

To construct such an independence conserving node, consider first the case where all variables are binary valued. The likelihood matrix for node $e$ is:

$$\begin{bmatrix} r & s \\ t & u \end{bmatrix} \stackrel{\text{def}}{=} p\{e^- | A B\} \text{ such that}$$

$$r \stackrel{\text{def}}{=} p\{e^- | A = a^+ \, B = b^+\},$$

$$s \stackrel{\text{def}}{=} p\{e^- | A = a^- \, B = b^+\} \text{ and so on.}$$

Taking expectation over $B$, the likelihood ratio seen by $A$, $p\{e^-|a^+\} / p\{e^-|a^-\}$, will be in the range between $r/s$ and $t/u$. It is evident that, if the likelihood ratios in each row are the same, then the likelihood ratio seen by the other parent, $A$, will be constant for any value of $B$. Thus the expected likelihood ratio for $A$ will be independent of the distribution of the other parent, node $B$. The same argument applies to the columns, and so to the relation of $B$ upon $A$.

This property generalizes to random variables with more than two states where each row in the likelihood matrix differs only by a ratio, so that the row space is of rank one. Using a well known result from linear algebra, the row rank equals the column rank, so the same argument applies to the columns' likelihood ratios. This suggests a way to construct such a matrix:

**Proposition 1:** Independence is preserved between direct predecessors $A$ and $B$ of a common successor node $E$ for one state of the evidence $e^-$, if the combined likelihood matrix is proportional to the "outer product" of the vectors for each individual likelihood:

$$p\{e^- | A B\} \propto p\{e^-|A\}p\{e^-|B\}.$$

This is shown by solving for $p\{A|B\,e^-\}$ for any $p\{A\}$, with Bayes' rule:

$$p\{A|B\,e^-\}$$
$$= \frac{p\{e^-|AB\}p\{A\}}{E_A[p\{e^-|AB\}p\{A\}]}$$
Substituting in the likelihood, and simplifying:
$$= \frac{p\{e^-|A\}p\{e^-|B\}p\{A\}}{E_A[p\{e^-|A\}p\{e^-|B\}p\{A\}]}$$
$$= \frac{p\{e^-\,A\}}{p\{e^-\}} = p\{A|e^-\}.$$

I will call this independence condition between predecessor nodes conditional on one state of the common evidence "conditional inter-causal independence," or CICI. This condition on the likelihood distribution serves as a qualification on the conditions of d-separation for specified states of evidence at $E$.

Since the likelihood matrix appears in both numerator and denominator of Bayes' rule, scaling the likelihood by a constant affects neither l.h.s. nor r.h.s. Thus in the binary case, where the likelihoods are $a = p\{e^-|a^+\}$, $b = p\{e^-|b^+\}$, the outer product of the two likelihood vectors with a scaling factor, $c$, is general form for a CICI relation matrix:

$$\begin{bmatrix} r & s \\ t & u \end{bmatrix} = \begin{bmatrix} abc & (1-a)bc \\ a(1-b)c & (1-a)(1-b)c \end{bmatrix}.$$

I will call this the "singular matrix" model. The independence constraint removes one degree of freedom, leaving the matrix to be specified with three parameters. For binary variables, this constraint is equivalent to the relation matrix having a determinant equal to zero. This follows from the proposition:

**Corollary 1:** The determinant of a likelihood matrix of binary valued random variables, $p\{e|AB\}$, of rank one equals zero. Thus $\det p\{e|AB\} = 0$ implies that $p\{A|Be\} = p\{A|e\}$. Multiplying out the determinant gives $\det p\{e|AB\} = ru - st$, the quantity referred to as "multiplicative synergy" by Henrion. [Henrion, Druzdzel 1990]

---

[1] W. Buntine has pointed out that this is also a well known property of the logistic distribution, which may be thought of as a continous version of the noisy-or.

This independence relation $p\{A|BE\} = p\{A|E\}$ holds for CICI nodes at both certainty for one value of $e = E$ as well as for complete ignorance of $E$. The next questions are 1) whether this independence is implied for all distributions $p\{E\}$, and conversely 2) whether there are necessarily states of $E$ for which CICI nodes do create conditional dependence. If 1) is true, CICI evidence nodes would be degenerate and serve no purpose.

To answer the first question we test if is it possible to have a relation matrix that is rank one at each state of the evidence. In that case the relation matrix would be factorable for every state of the evidence. In the binary case, this pair of constraints for both $E = e^+$ and $E = e^-$ can be shown, with some algebra, to imply that the likelihood ratios for one of the two parents must be constant and equal to one. This means that effectively there is no arc from that parent to the evidence. This independence is implied by a more general result of [Geiger and Heckerman 1990] about "transitive distributions" for which connectedness in graphical representations is equivalent to dependence among the distributions. Strictly positive binary distributions are one case of transitive distributions.

Now the converse, to show when the likelihood is factorable at one state of evidence it creates dependencies among parents at others. Let the evidence be a binary node, factorable at $E = e^-$. Then by Bayes rule, at the other state of the evidence:

$$\frac{p\{B|Ae^+\}}{p\{B\}} = \frac{p\{e^+|AB\}}{p\{e^+|A\}}$$
$$= \frac{(1 - p\{e^-|A\}p\{e^-|B\})}{1 - p\{e^-|A\}}$$

The right side cannot be factored into $A$ and $B$ factors, and is dependent upon $A$.

## 1.2 The noisy or

The noisy-or model is an example that illustrates the dependencies generated by CICI likelihoods:

**Proposition 2**: A "noisy-or" is a case of a CICI node. This can be shown by writing the noisy-or for evidence $e^+$ as

$$p\{e^+|AB\} = \begin{bmatrix} 1 - q_0q_1q_2 & 1 - q_0q_1 \\ 1 - q_0q_2 & 1 - q_0 \end{bmatrix},$$

where $q_i = 1 - p_i$, the reliability probabilities. It is evident that for evidence $e^-$, the likelihood matrix is a matrix of ones minus this. Calculating its determinant,

$$\det|1 - p\{e^+|AB\}| = \det p\{e^-|AB\} = 0.$$

With CICI nodes I will, by convention, label the evidence $e^-$ at which independence occurs.

The other way to build a CICI node is from the "singular matrix model," mentioned in the previous section, where the singular matrix represents the likelihood $p\{e^-|AB\}$. What is the relation between these two models? They both have three degrees of freedom. Equating and solving obtains $c = q_0$, $q_2 = b/(1-b)$, $q_1 = a/(1-a)$. Since all terms must be probabilities in the range of $(0,1)$, the noisy-or can be identified with the singular matrix model only when the singular matrix parameters are restricted to $0 < a, b < 1/2$. This is because the noisy-or model enforces a size ordering among matrix entries, the largest entry being in the upper left hand corner. There are three other cases, $0 < a < 1/2 \leq b < 1$, $0 < b < 1/2 \leq a < 1$ and $1/2 \leq a, b < 1$. These are equivalent to the noisy-or matrix with the row terms switched, the column terms switched, or both switched. These four generalizations cover the range of binary CICI relation nodes.

## 1.3 THE DEGREE OF INTER-CAUSAL EFFECT

We have seen that inter-causal independence among a node's parents depends upon the common node's evidence. In the binary case, forcing inter-causal independence at one state of the evidence precludes it from the other state. We have also seen that, in the binary case, the rank one condition for independence is easily tested by looking for a zero determinant of the likelihood matrix. The next question is, what does the value of a non-zero determinant indicate about the effect of $A$ upon $B$?

### 1.3.1 Qualitative effects

The value of this determinant varies from minus unity to plus unity as the relation between parents goes from extreme exclusion to extreme collaboration. At each extreme the parents $A$ and $B$ are deterministically dependent. Then either the parents are mutual exclusive, a condition already discussed, or they are forced to have identical distributions. To force identity between parents, the relation matrix becomes an identity matrix. For exclusion it is one minus this matrix— zeros on the diagonal and ones off-diagonal. Call these extremes "complete collaboration and "complete exclusion." Thus a relation matrix with complete collaboration for $e^+$ will have complete exclusion for $e^-$. These two matrices and their linear combinations are not CICI matrices, except for the trivial case of a constant matrix.

To be able to use the determinant measure—the multiplicative synergy—to characterize the relation between parents, I must first establish that the sign of this property of the likelihood matrix is invariant to Bayes' rule: The next theorem shows that the sign of the multiplicative synergy equals the sign of the CICI relation between parents not just for $p\{e|AB\}$, but for $p\{A|Be\}$, and all other permutations that may be generated by Bayes' rule.

**Lemma:** Multiplication of a likelihood matrix, $L(X,Y)_{ij}$ by any positive probability vector $v(X)_i$ does not change the sign of the likelihood's determinant.

To show this: Multiplication by a row vector variable is equivalent to a term-by-term multiplication of matrices where the vector is replicated to fill out the columns of its matrix. This, in turn, is equivalent to matrix multiplication where the vector values fill the diagonal of a matrix, with all other entries zero. Write this diagonal matrix derived from the vector as $d(v)_{ii}$. From linear algebra there is the result that the determinant of a product equals the product of each matrix's determinant, thus

$$\det d(v)_{ii} L(X,Y)_{ij} = \det d(v)_{ii} \det L(X,Y)_{ij}.$$

The determinant of the diagonal matrix is merely the product of terms along the diagonal, a number between zero and one. We can now show:

**Proposition 3:** Exclusion or collaboration (the sign of the multiplicative synergy) is given by the sign of the determinant of $p\{e|AB\}$ and is invariant to all permutations derivable by Bayes' rule of this likelihood matrix for a given conditioning.

Bayes' rule consists of multiplying the likelihood matrix by one probability vector, the prior, then dividing it by another, the pre-posterior. By the previous lemma, multiplication by the prior multiplies the likelihood's determinant by a positive number. Division by the pre-posterior likewise multiplies it by the reciprocal, another positive number. Both operations preserve the sign of the likelihood determinant. Note that since the conditioning of the likelihood must be preserved; $\det p\{e^+|AB\} > 0$ does not necessarily imply that $\det p\{a^+|EB\} > 0$.

### 1.3.2 Comparision to other measures of diagnostic and inter-causal relations

Inter-causality has been examined as a qualitative relation by Wellman. [Wellman 1988] In the tradition of non-numeric, automatic reasoning methods for planning, he has developed an abstraction of influence diagrams where each influence is described by its sign. These "qualitative probabilistic networks" can formulate decision tradeoffs by considering dominance relationships among alternatives. Such networks are constructed from two kinds of qualitative relations: the first, *qualitative influences*, describes the relation between two variables; the second is the relation between influences that he terms *qualitative synergy*, which corresponds to inter-causality. Here is his definition of synergy, in our notation: [p. 74]

**Definition:(Qualitative synergy)** Variables $A$ and $B$ are positively synergistic on $E$, written $Y^+(E|AB)$, or just $Y^+E$, if and only if, for every $x$, $a_1$, $a_2$, $b_1$, $b_2$, $e_0$, $a_1 \geq a_2$, $b_1 \geq b_2$ implies

$$p\{e_0|a_1 b_1 x\} - p\{e_0|a_2 b_1 x\}$$
$$\leq p\{e_0|a_1 b_2 x\} - p\{e_0|a_2 b_2 x\}.$$

Similarly in the last relation, substitute "$\geq$" for negatively synergistic and "$=$" for zero synergy.

Henrion has called this quantity "additive synergy" to distinguish it from the multiplicative synergy measure defined previously. In comparison to Wellman, our definition of "quantitative additive synergy" takes the liberty of assigning a value to Y whose sign corresponds to the sign of the synergy:

$$Ye^+ \stackrel{\text{def}}{=} Y(E = e^+|AB) = r + u - s - t.$$

Wellman does recognize in his examples the implied inter-causal relation between $A$ and $B$ due to the synergistic properties of the likelihood. As a further distinction, Wellman takes pains to extend his definition over all states of conditioning variables $x$, which he calls the *context*. This would be useless for our quantitative definition; however it serves his purpose of determining dominance relations. Unlike his definition however, I define a $Ye$ for each conditioning of $E$ in the likelihood matrix. Since qualitative synergy is derived from a stochastic dominance relation on continuous variables, to apply it to the case of binary variables he introduces a sign ordering convention such that $e^+ > e^-$. In my framework, his definition is equivalent to just the case where $E = e^+$. As such, the manner in which this relation depends upon the evidential support at $E$ is not developed in his examples.

### 1.3.3 Relation between additive and multiplicative synergies

As seen, for purposes of characterizing the effects between inter-causal nodes, I have modified definitions of synergy to be conditional on the states of binary variables. The next part develops a constraint among determinants (multiplicative synergy measures) of the same relation matrix with different states of binary evidence.

**Proposition 4:** Additive synergy equals the sum of the determinant measures, $\det e$, for both states of evidence. Expressed as a formula,

$$Ye^+ = \det e^+ - \det e^-,$$

where $\det e^+$ is defined to equal determinant $|p\{E = e^+|AB\}|$, and likewise $\det e^-$ to equal the determinant $|p\{E = e^-|AB\}|$. Further, $Y$ changes sign when the state of evidence is negated. To demonstrate, since $E$ is a binary variable,

$$\begin{aligned} Ye^- &= Y(E = e^-|AB) \\ &= 1 - r + 1 - u - (1 - s) - (1 - t) \\ &= s + t - r - u = -Ye^+. \end{aligned}$$

To see the relation between multiplicative and additive synergies, write out

$$\begin{aligned} \det e^- &= (1 - r)(1 - u) - (1 - s)(1 - t) \\ &= s + t - r - u + ru - st \\ &= Ye^- + \det e^+, \\ \text{or} \quad Ye^+ &= \det e^+ - \det e^- \end{aligned}$$

**Proposition 5:** Multiplicative and additive synergy are equal for CICI relation matrices. If one of the states of evidence forces independence (e.g., is CICI) then the determinant for that state disappears. Thus for CICI nodes the relation between additive and multiplicative synergy is: $\det e^+ = Ye^+$, that is, both measures are equivalent.

The additive-multiplicative synergy relation makes it easy to show the following:

**Proposition 6:** Noisy-or matrices are exclusionary nodes for $E = e^+$.

Since $\det e^- = 0$, one can use the previous result to show $Ye^+ < 0$. See [Agosta 1991].

A typical situation expressed by a noisy-or is the relation between seeing cat prints in someone's house and inferring which kind of cat they have as a pet. The exclusionary property of noisy-or nodes is the essence of their ability to "explain away" one hypothesized cause as another cause becomes more likely. Thus upon seeing paw prints, one cause—a pet blue Persian—tends to exclude their being also a short haired red tabby in the house.[2] If we comb the house and find no paw prints, the explanations remain independent: we are no wiser about relative probabilities of the household's domestic animals, even though we may justifiably tend to doubt they own a pet.

How would collaborative nodes, e.g. $Ye^+ > 0$ nodes, be constructed? Recall the result in Linear Algebra that switching a pair of rows or columns of a matrix switches the sign of a matrix's determinant. Thus they can be built from exclusionary nodes by switching the off-diagonal and on-diagonal elements.

---

[2] For the model to apply strictly, there should be no relation between lovers of different kinds of cats; that is, being a Persian owner should not, in itself, make the household more or less likely to own a short haired tabby. (This example is inspired by [M. Henrion, 1990].)

### 1.3.4 The range of inter-causal dependency

How can the dependency be described quantitatively? This inter-causal dependency is not just a consequence of the diagnostic dependencies between the parents, $A$ and $B$, and the evidence; rather it may be thought of as the relation between these dependencies. At the extremes of complete inter-causal dependency, the individual (marginal) likelihoods $p\{E|A\}$ and $p\{E|B\}$ are completely determined by the marginals of the other predecessor: there is no additional freedom in the diagnostic relation between hypothesis and evidence. In comparison, when $A$ and $B$ are inter-causally conditionally independent, the diagnostic support between hypothesis and evidence for each can be specified independently.

As a consequence of proposition 3, there is a qualitative correspondence, where the sign of the determinant of likelihood matrix terms $p\{e|AB\}$ corresponds to the sign of the induced dependency of $p\{A|B\}$. Their quantitative relation is not as obvious. Note that unlike the determinant, $\det e^+$, $p\{A|B\}$ is homogeneous of zeroth order in the likelihood terms. That is, scaling the entries in the likelihood matrix does not change the dependence among parents, as can be seen from the following version of Bayes' rule:

$$p\{B|Ae\} = \frac{p\{e|AB\}p\{B\}}{E_B[p\{e|AB\}]}$$

This means that multiplying all terms of $p\{e|AB\}$ by a constant changes the value of the determinant but leaves $p\{B|Ae\}$ unchanged, destroying the one to one correspondence between the multiplicative synergy and any quantitative characterization of the intercausal relation $p\{e|AB\}$.

To explore the quantitative relation, the next section shows the construction of the algebraic solution for one parent's belief as a function of the rest of the clique's nodes.

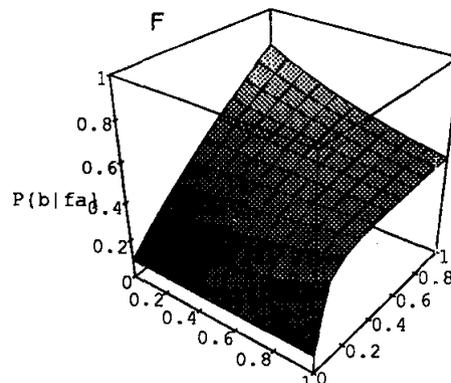

Figure 2: The noisy-or belief surface

## 2 CONSTRUCTIVE SOLUTION OF THE BINARY VARIABLE INTER-CAUSAL DEPENDENCY

By deriving the belief of one parent as a function of the probabilities of the other nodes in the clique, one may examine its quantitative behavior completely. This function is a three-dimensional surface, a marginal probability of one parent as a function of the support for their common evidence and the other parent. This "belief function surface" shows the combined effect on one parent of the diagnostic and inter-causal influences.

The solution technique used is similar to the clique potential methods. All the nodes are formed into one clique potential, $\Psi$, proportional to the joint for the state space of all nodes. The probability that we solve for is the posterior on $B$ as a function of the probabilities of other nodes in the network. To make precise the sensitivity of one node's marginal on other nodes' probabilities, think of the "knobs" to control the other nodes as pi ($\pi$) and lambda ($\lambda$) messages to the nodes; $\pi$ messages as the root nodes, and $\lambda$ messages to leaf nodes. These messages can be specified independently of each other, whereas in general the marginal probabilities of nodes cannot, since they are not independent. In this case there is one $\lambda$ message, to the evidence node, $E$, and two $\pi$ messages, one for each parent, of which we are mainly interested in the $\pi$ for the "other" parent, $A$.

The posterior on $B$ is a function of both parent priors, $\pi(a)$ and $\pi(b)$, the evidence likelihood matrix, $p\{E|AB\}$, and the "evidential support," or the $\lambda$ message that the evidence receives. To show the functional dependence, I write $p\{B|\pi(a)\pi(b)\lambda(e)\}$. It is important to distinguish this from $p\{B|AE\}$, which is a tabulation of probabilities at each combination of points in the state space, rather than a function of probabilities. The potential, $\Psi$, is a $2 \times 2 \times 2$ matrix, the product of all terms. To obtain the posterior on $B$, sum over all other variables, then normalize by the sum of all eight terms. These definitions are used for clarity:

$$a \stackrel{\text{def}}{=} \pi(A = a^+), b \stackrel{\text{def}}{=} \pi(B = b^+), f \stackrel{\text{def}}{=} \lambda(E = e^+),$$

so that,

$$\Psi(a,b,f) = \pi(A)\pi(B)\lambda(E)p\{E|AB\}.$$

Thus,

$$p\{B|\pi(a)\pi(b)\lambda(e)\} = \frac{\sum_{AE} \Psi(a,b,f)}{\sum_{ABE} \Psi(a,b,f)}$$

$$= \frac{\begin{array}{c}b[a(fr + (1-f)(1-r)) \\ + (1-a)(sf + (1-f)(1-s))], \\ (1-b)[a(ft + (1-f)(1-t)) \\ + (1-a)(uf + (1-f)(1-u))]\end{array}}{\sum_{ABE} \Psi(a,b,f)}$$

The numerator is a two-valued vector for $b^+$ and $b^-$. It is normalized by the denominator, which is precisely the sum of the two terms in the numerator. Figure two shows a graph of this "belief surface" as a function of $f$ and $a$, for $\pi(b) = 1/2$. The values from this example are for a symmetric noisy-or likelihood matrix. The conditional independence of the parent node probabilities is evident by the constant value of the function for all values of $a$ at both $f = 0$; that is, $e^- = E$ and $f = 1/2$, complete ignorance of $E$. The exclusionary property is evident along the edge $f = 1$, where $B$ is inversely related to $a$. The graph may be thought of as a combination of the diagnostic relation, where decreasing $f$ increases the belief in both parents, together with an inter-causal exclusionary relation when $e^+ = E$. The inter-causal relation is slight when $1/2 \leq f \leq 0$ since at both extremes of this interval the inter-causal dependence disappears.

The exclusive relation between the beliefs of $A$ and $B$ are described by the $f = \lambda(e) = 1$ edge of the belief function surface, shown here in figure three:

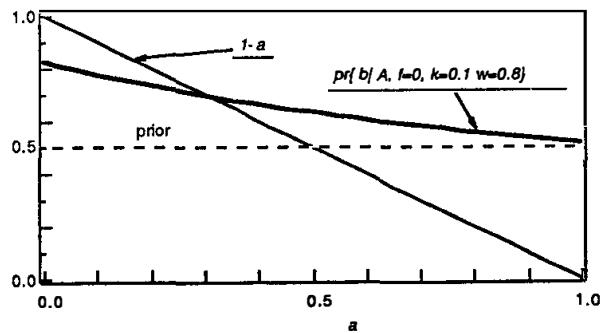

Figure 3: Partial exclusion of a CICI relation, bounded below by its prior. The other line shows the complete exclusion relation.

The two beliefs move in opposite directions, with $p\{B|\pi(a)\pi(b)\lambda(e) = 1\}$ having a maximum approximately at $w$ and a minimum no lower than $\pi(b)$. Previously I mentioned a relation matrix for complete exclusion, which forced $p\{B|e\} + p\{A|e\} = 1$. In comparison to this *partial exclusion* at $f = 1$, the complete exclusion probability, $p\{B|e\}$, descends to zero from unity. The outstanding difference between complete exclusion and that generated by a CICI node is this lower bound that prevents $B$'s belief from ever being driven below its value without the CICI node. Thus "CICI partial exclusion" cannot defeat other support for a node's belief.

The degree of inter-causal exclusion is limited by the diagnostic effect when their combination operates in opposite directions. One way to think of this is that the positive effect of the diagnostic support dominates the negative effect of exclusion. In the "cat household" example, if we see a blue Persian after having seen paw prints, our belief in the presence of a short haired red tabby cannot be less than our prior belief about the tabby. To formalize this property of CICI nodes:

**Proposition 7:** For $p\{e^+|AB\}$ in the form of a noisy-or, $p\{B|A\,e^+\} > \pi(b)$ for all values of $\pi(a)$. In terms of the graph in figure two, this constrains the $e^+$ half of the belief function surface to lie above the prior value, and the $e^-$ half to lie below. The surface intersects the $\pi(b)$ valued horizontal plane only along the $E = 1/2$ line.

To show

$$p\{B|\pi(a) = 1\pi(b)\lambda(e) = 1\} > \pi(b),$$

write it out in functional form;

$$\frac{br}{br + (1-b)t} > b,$$

which reduces to $r > t$, an assumption of the noisy-or.

In a corresponding manner the size ordering relative to $B$ of the other three vertices of the belief surface can be demonstrated. Each vertex value is an increasing function of the prior on $B$ and the ratio of a pair of elements in the likelihood matrix. For both $p\{B|e^-a^+\}$ and $p\{B|e^-a^-\}$, the "independent edge" vertices, the ratios are equal: $(1-r)/(1-t) = (1-s)/(1-u)$. This is just a restatement of the $\det e^- = 0$ condition.

The "independent edge" value $p\{B|e^-\}$ and the "positive exclusion" value $p\{B|a^-e^+\}$, the two extreme values of the surface, describe the surface completely, and have physical significance in the model. I will use them to effectively factor the relation into a two parameter model of the likelihood, in the "factored" form of a symmetric noisy-or:

$$p\{e^-|AB\} = \begin{bmatrix} k^2w & kw \\ kw & w \end{bmatrix} \text{ for } 0 < k < 1, 0 < w < 1.$$

With the belief surface we can describe qualitatively both parameters' effects. As $w$ increases, the "positive exclusion" vertex, $p\{B|a^-e^+\}$, increases also. As $k$ decreases, the vertex probabilities become more extreme. At the same time, the "negative exclusion" vertex approaches $\pi(b)$. This is also true for non-symmetric noisy-or's, thus the degree of freedom that was lost to the symmetry assumption has only marginal effect on the surface shape. Further, when $\pi(b)$ approaches either zero or one it pulls the whole surface with it, for instance as $\pi(b) \to 1$ then $p\{B|AE\} \to 1$.

To derive the vertex values in the limit of small $k$ and large $w$, approximate the values by first order expansions in $k$ and $w$. First this lemma, by which one may approximate rational functions whose numerator and denominator differ by a "small" amount:

**Lemma:** Since

$$\frac{1}{1-z} = 1 + z + \frac{z^2}{1-z}$$

this approximation holds:

$$\frac{1}{1-z} = 1 + z + O(z^2) \geq 1 + z, \text{ for } z \text{ small}.$$

With this formula the best linear approximation to a rational polynomial is obtained without the need to write out the derivative.

**Proposition 8:** The "independent edge" probability $p\{B|e^-\}$ at $\pi(b) = 1/2$ is independent of $w$ and equals $k/(1+k)$. This follows exactly since

$$p\{B|e^-\} = \frac{bk}{1+b(k-1)}.$$

Further, $k$ sets an upper bound for this probability, since it follows that for all $k$ and $b$,

$$p\{B|e^-\} < k.$$

**Proposition 9:** The "negative exclusion" corner $p\{B|a^-e^-\}$ approaches $B$ from above, such that

$$p\{B|a^-e^-\} \geq b[1 + kw(1-b)].$$

Since the inequality is bounded by $O(z^2)$, this probability approaches $b$, linearly in $k$, as $k$ approaches 0. When $k$ is small $p\{B|a^-e^-\}$ is well approximated by $b$.

**Proposition 10:** The "positive exclusion" probability $p\{B|a^-e^+\}$ is bounded below to $O(z^2)$ such that

$$p\{B|a^-e^+\} > 1 - \frac{(1-b)(1-w)}{b(1-kw)}.$$

Further, when $k$ is small and $b$ is near $1/2$, this limit is approximately equal to $w$.

To summarize, it is a good approximation that the belief surface, and hence any CICI distribution, can be specified by limits to the minimum and maximum values of the surface, which imply the conditional probabilities of the parent nodes at different states of evidence. These probabilities lead directly to estimates of the symmetric CICI likelihood parameters; $k$ approaching the "independent edge" conditional probability, and $w$ approaching the "positive exclusion" conditional probability. The remaining vertex, the "negative exclusion" conditional probability closely approximates the parent's prior. The error in the approximation is second order in $k$ and $1 - w$, and the approximation becomes exact as $k \to 0$ and $w \to 1$.

## 3  DISCUSSION

A major finding of this paper is that the CICI effect of evidence is secondary to its diagnostic effects. Thus the relative effect between hypotheses—call it the observed exclusion—is also a consequence of the degree of direct support for the hypotheses as much as it is affected by the partial exclusion controled by the noisy-or parameter, $w$. The more that two related hypotheses have direct support, the less that secondary inter-causal effects appear. Thus the refutational effect of $w$ on a hypothesis due to conflicting hypotheses decreases as other support for the hypothesis increases.

This paper has explored the properties of CICI evidence nodes. The properties are two: First, when it is certain that the evidence is absent, e.g. at $e^-$, the CICI node leaves dependencies among the hypothesis set unchanged. For hypotheses that are otherwise independent, this reduces the connectivity of the network, and thus simplifies the complexity of the probability updating algorithm. Secondly, at the other extreme when the evidence, $e^+$, is present, the CICI node generates partial exclusion (or collaboration) among the set, in the sense that the exclusion can not decrease other evidential support, only increase support in the lack of other evidence.

There are several consequences of building a network of nodes with these properties. First, the conditional independence property implies the exclusion property, so we either accept both, or neither. It is a general property of common evidence nodes, not only CICI nodes, that shared evidence generates dependencies among hypotheses; and we have seen that we cannot have independence among the existence of hypotheses for all states of evidence. As a consequence, it is probabilistically inconsistent to treat common evidence separately, inferring each hypothesis independently. This can be summed up in the phrase "ambiguity implies conflict," meaning that alternate, competing explanations must probabilistically exclude each other. Conversely, they could also be collaborating explanations that become coupled by common evidence. What is not possible is for two perfectly good explanations of a common effect to be probabilistically independent of each other for all states of the evidence.

Multiple parent nodes are the elements from which to build networks of multiply connected hypotheses. This technique is similar to other "constraint propagation networks" of hypotheses where typically inter-hypothesis constraints are expressed without intervening nodes. Constraint networks typically can propagate a small change through all nodes in a network, because of their similar properties to sets of simultaneous equations. In comparison, inter-causal constraints tend to have a quickly attenuated effect among chains of nodes, since the percent change diminishes from a node to its neighbor. Inter-causal constraints are best thought of as resulting in a secondary set of effects that tend to increase the discrimination of diagnostic inference among hypotheses.

### Acknowledgements

My grateful acknowledgements to Tom Binford, Max Henrion, Harold Lehmann, Gregory Provan, Ross Shachter and Mike Wellman for their comments and suggestions. Also to Wally Mann and Margaret Miller for help with the figures.